\documentclass[10pt]{article}

\usepackage{amsmath,amsthm,amsfonts,mathrsfs,anysize,fancyhdr,epsfig,mdframed, float,graphicx,umoline,dsfont}

\usepackage{chngcntr}
\usepackage{apptools}

\usepackage{amsmath}                                                        
\usepackage{graphicx}                                                       
\usepackage{subfigure}
\usepackage{enumitem}                                                    
\usepackage{hyperref}
\usepackage{amssymb}                                                        
\usepackage[mathscr]{eucal}                                             
\usepackage{chngcntr}
\counterwithin{table}{section}                                                                                   
\counterwithin{figure}{section}                                                                                   
\usepackage{cancel}                                                             
\usepackage[normalem]{ulem}                                                                 
\usepackage{pstricks}
\usepackage{rotating}
\usepackage{lscape}
\usepackage[paperwidth=8.5in,paperheight=11in,top=1.25in, bottom=1.25in, left=1.00in, right=1.00in]{geometry}
\usepackage{mathtools}                                                      
\mathtoolsset{showonlyrefs=true}                                    
\usepackage{fixltx2e,amsmath}                                           
\MakeRobust{\eqref}

\providecommand{\keywords}[1]{\textbf{Keywords:} #1}
\usepackage{mathdots}
\usepackage{amsthm}                                                             
\allowdisplaybreaks                                                             
                                                               
\newtheorem{remark}{Remark}
\begin{document}

\title{Conditional time series forecasting with convolutional neural networks}
\author{Anastasia Borovykh\thanks{Dipartimento di Matematica, Universit\`a di Bologna, Bologna, Italy.
\textbf{e-mail}: anastasia.borovykh2@unibo.it} 
\and Sander Bohte \thanks{Centrum Wiskunde \& Informatica, Amsterdam, The Netherlands. \textbf{e-mail}:s.m.bohte@cwi.nl} \and Cornelis W. Oosterlee \thanks{Centrum Wiskunde \& Informatica, Amsterdam, The Netherlands. \textbf{e-mail:} c.w.oosterlee@cwi.nl}\thanks{Delft University of Technology, Delft, The Netherlands}}
\date{This version: \today}
 
 \maketitle
 
\begin{abstract}
We present a method for conditional time series forecasting based on an adaptation of the recent deep convolutional WaveNet architecture. The proposed network contains stacks of dilated convolutions that allow it to access a broad range of history when forecasting, a ReLU activation function and conditioning is performed by applying multiple convolutional filters in parallel to separate time series which allows for the fast processing of data and the exploitation of the correlation structure between the multivariate time series. We test and analyze the performance of the convolutional network both unconditionally as well as conditionally for financial time series forecasting using the S\&P500, the volatility index, the CBOE interest rate and several exchange rates and extensively compare it to the performance of the well-known autoregressive model and a long-short term memory network. We show that a convolutional network is well-suited for regression-type problems and is able to effectively learn dependencies in and between the series without the need for long historical time series, is a time-efficient and easy to implement alternative to recurrent-type networks and tends to outperform linear and recurrent models.
\end{abstract}

\section{Introduction}
Forecasting financial time series using past observations has been a topic of significant interest for obvious reasons. It is well known that while temporal relationships in the data exist, they are difficult to analyze and predict accurately due to the non-linear trends, heavy tails and noise present in the series \cite{cont01}. In developing models for forecasting financial data it is desirable that these will be both able to learn non-linear dependencies in the data as well as have a high noise resistance. Traditional autoregressive models such as VAR and ARMA \cite{hamilton94} fail to capture non-linear patterns. Feedforward neural networks have been a popular way of learning the dependencies in the data as these allow to learn non-linearities without the need of specifying a particular model form in advance, see \cite{zhang98} or \cite{chakraborty92}. Hybrid approaches using neural networks and econometric models have also been proposed, e.g. \cite{zhang03}. One downside of classical feedforward neural networks is that a large sample size of data is required to obtain a stable forecasting result.

The main focus of this paper is on multivariate time series forecasting, specifically financial time series. In particular, we forecast time series conditional on other, related series. Financial time series are known to both have a high noise component as well as to be of limited duration -- even when available, the use of long histories of stock prices can be difficult due to the changing financial environment. At the same time, many different, but strongly correlated financial time series exist. Here, we aim to exploit multivariate forecasting using the notion of conditioning to reduce the noisiness in short duration series. Effectively, we use multiple financial time series as input in a neural network, thus conditioning the forecast of a time series on both its own history as well as that of multiple other time series. Training a model on multiple stock series allows the network to exploit the correlation structure between these series so that the network can learn the market dynamics in shorter sequences of data. As shown by e.g. \cite{zheng16} for classification, using multiple conditional time series as inputs can improve both the robustness and forecast quality of the model by learning long-term temporal dependencies in between series. 

A convolutional neural network (CNNs), see \cite{lecun98}, is a biologically-inspired type of deep neural network (DNN) that has recently gained popularity due to its success in classification problems (e.g. image recognition \cite{krizhevsky12} or time series classification \cite{wang16}). The CNN consists of a sequence of convolutional layers, the output of which is connected only to local regions in the input. This is achieved by sliding a filter, or weight matrix, over the input and at each point computing the dot product between the two (i.e. a convolution between the input and filter). This structure allows the model to learn filters that are able to recognize specific patterns in the input data. Recent advances in CNNs for time series forecasting include \cite{mittelman15} where the authors propose an undecimated convolutional network for time series modelling based on the undecimated wavelet transform and \cite{binkowski17} in which the authors propose to use an autoregressive-type weighting system for forecasting financial time series, where the weights are allowed to be data-dependent by learning them through a CNN. In general literature on financial time series forecasting with convolutional architectures is still scarce, as these types of networks are much more commonly applied in classification problems. Intuitively, the idea of applying CNNs to time series forecasting would be to learn filters that represent certain repeating patterns in the series and use these to forecast the future values. Due to the layered structure of CNNs, they might work well on noisy series, by discarding in each subsequent layer the noise and extracting only the meaningful patterns, in this way bearing similarities to neural networks which use wavelet transformed time series (i.e. a split in high- and low-frequency components) as input, see e.g. \cite{aussem97}, \cite{lahmiri14}.

Currently, recurrent neural networks (RNNs), and in particular the long-short term memory unit (LSTM) \cite{hochreiter97}, \cite{chung14}, are the state-of-the-art in time series forecasting, see also \cite{hsu17} and in particular \cite{fisher17} for financial forecasting results. The efficiency of these networks can be explained by the recurrent connections that allow the network to access the entire history of previous time series values. Alternatively one might employ a convolutional neural network with multiple layers of dilated convolutions \cite{yu15}. The dilated convolutions, in which the filter is applied by skipping certain elements in the input, allow for the receptive field of the network to grow exponentially, hereby allowing the network to, similar to the RNN, access a broad range of history. The advantage of the CNN over the recurrent-type network is that due to the convolutional structure of the network, the number of trainable weights is small, resulting in a much more efficient training and predicting. 

Motivated by \cite{vanoord16a} in which the authors compare the performance of the PixelCNN to the PixelRNN \cite{vanoord16b}, a network used for image generation, in this paper, we aim to investigate the performance of the convolutional neural network compared to that of autoregressive and recurrent models on forecasting noisy, financial time series. The CNN we employ is a network inspired by the convolutional WaveNet model from \cite{vanoord16} first developed for audio forecasting, whose structure we simplify and optimize for multivariate time series forecasting. Our network focusses on learning long-term relationships in and between multivariate, noisy time series. Similar to \cite{vanoord16} it makes use of the dilated convolutions, however these convolutions are applied with parametrized skip connections \cite{he15a} from both the input time series as well as the time series we condition on, in this way learning long- and short-term \emph{inter}dependencies in an efficient manner. Furthermore, the gated activation function from the original WaveNet model is replaced by a rectified linear unit (ReLU), simplifying the model and reducing the training time. 

This paper consists of several main contributions. First of all, we present a CNN inspired by the WaveNet model, with a structure that is simplified and optimized for time series forecasting, i.e. using a ReLU activation and a novel and more optimal way of conditioning with parametrized skip connections. Second, knowing the strong performance of CNNs on classification problems, our work is --to the best of our knowledge-- the first to show that they can be applied successfully to forecasting financial time series of limited length. By conducting an extensive analysis of the WaveNet model and comparing the performance to that of an LSTM, the current state-of-the-art in forecasting, and an autoregressive model popular in econometrics our paper shows that the WaveNet model is a time-efficient and easy to implement alternative to recurrent-type networks and tends to outperform the linear and recurrent models. Lastly, we show using examples on artificial time series as well as the S\&P500, VIX, CBOE interest rate and five exchange rates that the efficient way of conditioning in the WaveNet model enables one to extract temporal relationships in between time series improving the forecast, while at the same time limiting the requirement for a long historical price series and reducing the noise, since it allows one to exploit the correlations in between related time series. As a whole, we show that convolutional networks can be a much simpler and easier to train alternative to recurrent networks while achieving an at least as good or better accuracy on non-linear, noisy forecasting tasks.

\section{The model}
In this section we start with a review of neural networks and convolutional neural networks. Then we introduce the particular convolutional network structure that will be used for time series forecasting.
\subsection{Background}
\subsubsection{Feedforward neural networks}
A basic feedforward neural network consists of $L$ layers with $M_l$ hidden nodes in each layer $l=1,...,L$. Suppose we are given as input $x(1),...,x(t)$ and we want to use the multi-layer neural network to output the forecasted value at the next time step $\hat x(t+1)$. In the first layer we construct $M_1$ linear combinations of the input variables in the form
\begin{align}
a^1(i) = \sum_{j=1}^t w^1(i,j)x(j) + b^1(i),\;\;\textnormal{for }i=1,...,M_1,
\end{align}
where $w^1 \in\mathbb{R}^{M_1\times t}$ are referred to as the weights and $b^1\in\mathbb{R}^{M_1}$ as the biases. Each of the outputs $a^1(i)$, $i=1,...,M_1$ are then transformed using a differentiable, nonlinear activation function $h(\cdot)$ to give
\begin{align}
f^1(i) = h(a^1(i)), \;\;\textnormal{for }i=1,...,M_1.
\end{align}
The nonlinear function enables the model to learn nonlinear relations between the data points. In every subsequent layer $l=2,...,L-1$, the outputs from the previous layer $f^{l-1}$ are again linearly combined and passed through the nonlinearity
\begin{align}
f^l(i) = h\left(\sum_{j=0}^{M_{l-1}} w^l(i,j)f^{l-1}(j) + b^l(j)\right),\;\;\textnormal{for }i=1,...,M_1,
\end{align}
with $w^l\in\mathbb{R}^{M_{l}\times M_{l-1}}$ and $b^l\in\mathbb{R}^M_{l}$. In the final layer $l=L$ of the neural network the forecasted value $\hat x(t+1)$ is computed using
\begin{align}
\hat x(t+1) = h\left(\sum_{j=0}^{M_{L-1}} w^L(j)f^{L-1}(j) + b^L\right),
\end{align}
with $w^l\in\mathbb{R}^{1\times M_{l-1}}$ and $b^l\in\mathbb{R}$. In a neural network, every node is thus connected to every node in adjacent layers, see Figure \ref{nnvscnn}. 

\subsubsection{Convolutions}
A discrete convolution of two one-dimensional signals $f$ and $g$, written as $f * g$, is defined as
\begin{align}
(f*g)(i) &= \sum_{j=-\infty}^{\infty} f(j)g(i-j),
\end{align}
where depending on the definition of the convolution, nonexistent samples in the input may be defined to have values of zero, often referred to as zero padding, or computing the product only at the points where samples exist in both signals.
Note that a convolution is commutative, i.e. $(f*g)=(g*f)$. If the signals are finite, the infinite convolution may be truncated. In other words, suppose $f=[f(0),...,f(N-1)]$ and $g = [g(0),...,g(M-1)]$, the convolution of the two is given by 
\begin{align}
(f*g)(i) = \sum_{j=0}^{M-1}f(j)g(i-j).
\end{align}
The size of the convolution output depends on the way undefined samples are handled. If a certain amount of the undefined samples is set to zero this is referred to as zero padding. If we do not apply zero padding the output has size $N-M+1$ (so that $i=0,...,N-M$), while padding with $p$ zeros at both sides of the input signal $f$ results in an output of size $N-M+2p+1$. The zero padding thus allows one to control the output size of the convolution, adjusting it to be either decreasing, the same, or increasing with respect to the input size. A convolution at point $i$ is thus computed by shifting the signal $g$ over the input $f$ along $j$ and computing the weighted sum of the two.

\subsubsection{Convolutional neural networks}
\begin{figure}[h!]
\centering
\includegraphics[width=0.8 \textwidth]{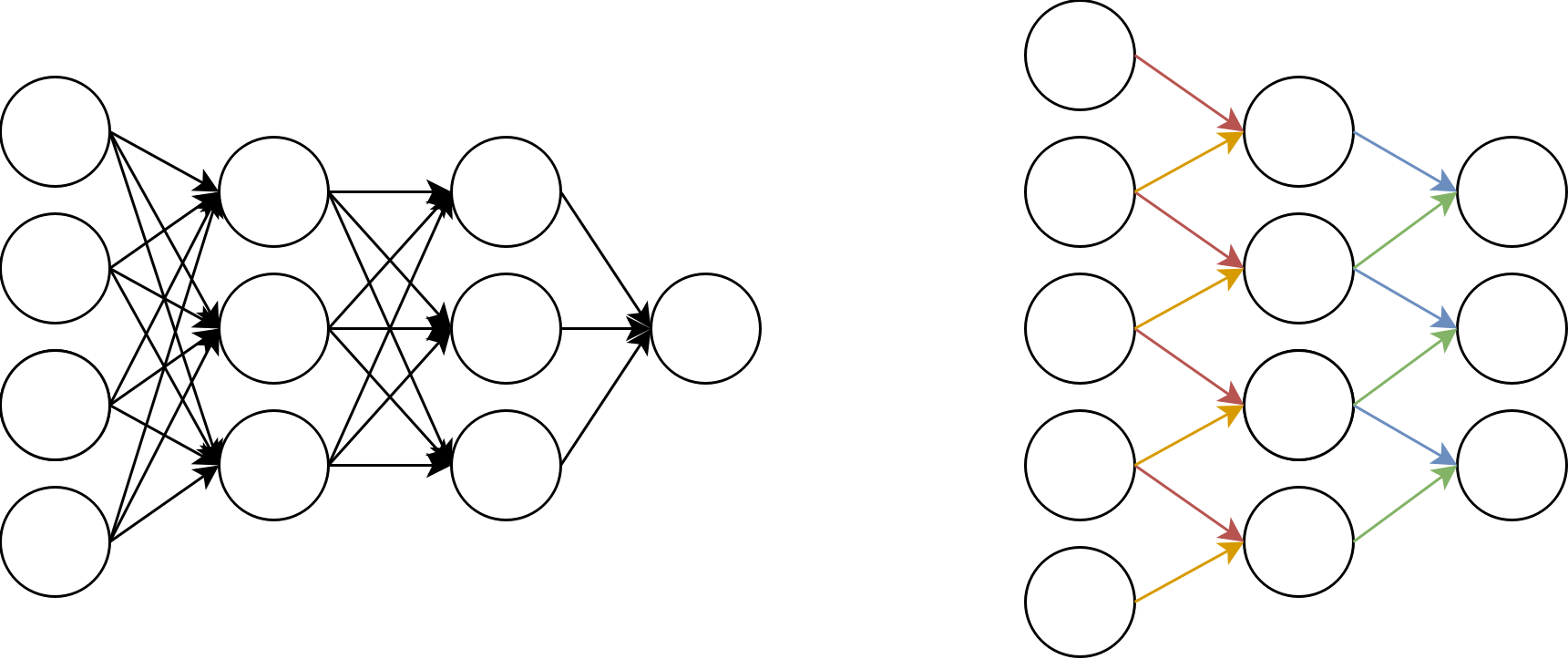}\hspace{0.5cm}       
\caption{A feedforward neural network with three layers (L) vs. a convolutional neural network with two layers and filter size $1\times 2$, so that the receptive field of each node consists of two input neurons from the previous layer and weights are shared across the layers, indicated by the identical colors (R).}\label{nnvscnn}
\end{figure}

Convolutional neural networks were developed with the idea of local connectivity. Each node is connected only to a local region in the input, see Figure \ref{nnvscnn}. The spatial extent of this connectivity is referred to as the receptive field of the node. The local connectivity is achieved by replacing the weighted sums from the neural network with convolutions. In each layer of the convolutional neural network, the input is convolved with the weight matrix (also called the filter) to create a feature map. In other words, the weight matrix slides over the input and computes the dot product between the input and the weight matrix. Note that as opposed to regular neural networks, all the values in the output feature map share the same weights. This means that all the nodes in the output detect exactly the same pattern. The local connectivity and shared weights aspect of CNNs reduces the total number of learnable parameters resulting in more efficient training. The intuition behind a convolutional neural network is thus to learn in each layer a weight matrix that will be able to extract the necessary, translation-invariant features from the input. 

The input to a convolutional layer is usually taken to be three-dimensional: the height, weight and number of channels. In the first layer this input is convolved with a set of $M_1$ three-dimensional filters applied over all the input channels (in other words, the third dimension of the filter map is always equal to the number of channels in the input) to create the feature output map. Consider now a one-dimensional input $x = (x_t)_{t=0}^{N-1}$ of size $N$ with no zero padding. The output feature map from the first layer is then given by convolving each filter $w^1_h$ for $h=1,...,M_1$ with the input:
\begin{align}
a^1(i,h) = (w^1_h * x)(i) = \sum_{j=-\infty}^\infty w^1_h(j)x(i-j),
\end{align}
where $w^1_h\in\mathbb{R}^{1\times k\times 1}$ and $a^1\in\mathbb{R}^{1\times N-k+1\times M_1}$. Note that since the number of input channels in this case is one, the weight matrix also has only one channel. Similar to the feedforward neural network, this output is then passed through the non-linearity $h(\cdot)$ to give $f^1 = h(a^1)$. 

In each subsequent layer $l=2,...,L$ the input feature map, $f^{l-1}\in \mathbb{R}^{1\times N_{l-1}\times M_{l-1}}$, where $1\times N_{l-1}\times M_{l-1}$ is the size of the output filter map from the previous convolution with $N_{l-1} = N_{l-2}-k+1$, is convolved with a set of $M_{l}$ filters $w^l_h\in\mathbb{R}^{1\times k\times M_{l-1}}$, $h=1,...,M_{l}$, to create a feature map $a^l\in\mathbb{R}^{1\times N_l \times M_{l}}$:
\begin{align}
a^l(i,h) = (w^l_h * f^{l-1})(i) = \sum_{j=-\infty}^\infty\sum_{m=1}^{M_{l-1}}w_h^l(j,m)f^{l-1}(i-j,m).
\end{align}
The output of this is then passed through the non-linearity to give $f^l = h(a^l)$. The filter size parameter $k$ thus controls the receptive field of each output node. Without zero padding, the convolution output in every layer has width $N_l = N_{l-1}-k+1$ for $l=1,..,L$. Since all the elements in a feature map share the same weights this allows for features to be detected in a time-invariant manner, while at the same time it reduces the number of trainable parameters. The output of the network after $L$ convolutional layers will be the matrix $f^L$, the size of which depends on the filter size and number of filters used in the final layer. Depending on what we want our model to learn, the weights in the model are trained to minimize the error between the output from the network $f^L$ and the true output we are interested in. 

\subsection{Structure}
Consider a one-dimensional time series $x = (x_t)_{t=0}^{N-1}$. Given a model with parameter values $\theta$, the task for a predictor is to output the next value $\hat x(t+1)$ conditional on the series' history, $x(0),...,x(t)$. This can be done by maximizing the likelihood function 
\begin{equation}
p(x|\theta) = \prod_{t=0}^{N-1}p(x(t+1)|x(0),...,x(t),\theta).
\end{equation}
To learn this likelihood function, we present a convolutional neural network in the form of the WaveNet architecture \cite{vanoord16} augmented with a number of recent architectural improvements for neural networks such that the architecture can be applied successfully to time series prediction. 

\begin{figure}[h!]
\centering
\includegraphics[width=0.6 \textwidth]{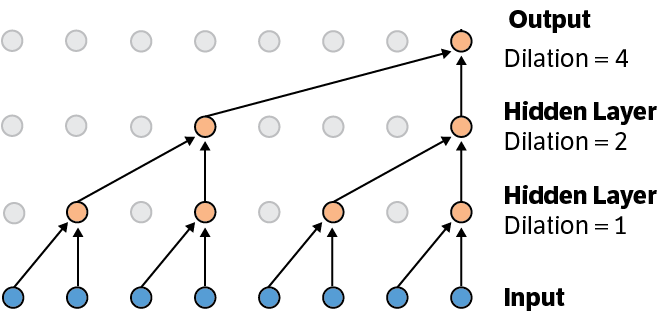}   
\caption{A dilated convolutional neural network with three layers.}\label{figdc}
\end{figure}

Time series often display long-term correlations, so to enable the network to learn these long-term dependencies we use stacked layers of dilated convolutions. As introduced in \cite{yu15}, a dilated convolution outputs a stack of $M_{l}$ feature maps given by
\begin{equation}
(w^l_h *_d f^{l-1})(i) = \sum_{j=-\infty}^{\infty}\sum_{m=1}^{M_{l-1}}w^l_h(j,m)f^{l-1}(i-d\cdot j,m),
\end{equation}
where $d$ is the dilation factor and $M_l$ the number of channels. In other words, in a dilated convolution the filter is applied to every $d$th element in the input vector, allowing the model to efficiently learn connections between far-apart data points. We use an architecture similar to \cite{yu15} and \cite{vanoord16} with $L$ layers of dilated convolutions $l=1,...,L$ and with the dilations increasing by a factor of two: $d \in [2^0,2^1,...,2^{L-1}]$. The filters $w$ are chosen to be of size $1\times k := 1\times 2$. An example of a three-layer dilated convolutional network is shown in Figure \ref{figdc}. Using the dilated convolutions instead of regular ones allows the output $y$ to be influenced by more nodes in the input. The input of the network is given by the time series $x=(x_t)_{t=0}^{N-1}$. In each subsequent layer we apply the dilated convolution, followed by a non-linearity, giving the output feature maps $f^l$, $l=1,...,L$. These $L$ layers of dilated convolutions are then followed by a $1\times 1$ convolution which reduces the number of channels back to one, so that the model outputs a one-dimensional vector. Since we are interested in forecasting the subsequent values of the time series, we will train the model so that this output is the forecasted time series $\hat x = (\hat x_t)_{t=0}^{N-1}$. 

The receptive field of a neuron was defined as the set of elements in its input that modifies the output value of that neuron. Now, we define the receptive field $r$ of the model to be the number of neurons in the input in the first layer, i.e. the time series, that can modify the output in the final layer, i.e. the forecasted time series. This then depends on the number of layers $L$ and the filter size $k$, and is given by
\begin{align}
r:= 2^{L-1}k.
\end{align} 
In Figure \ref{figdc}, the receptive field is given by $r=8$, one output value is influenced by eight input neurons. 

As mentioned before, sometimes it is convenient to pad the input with zeros around the border. The size of this zero-padding then controls the size of the output. In our case, to not violate the adaptability constraint on $x$, we want to make sure that the receptive field of the network when predicting $x(t+1)$ contains only $x(0),...,x(t)$. To do this we use causal convolutions, where the word causal indicates that the convolution output should not depend on future inputs. In time series this is equivalent to padding the input with a vector of zeros of the size of the receptive field, so that the input is given by 
\begin{align}
[0,...,0,x(0),...,x(N-1)]\in \mathbb{R}^{N+r},
\end{align}
 and the output of the $L$-layer Wavenet is 
 \begin{align}
 [\hat x(0),...,\hat x(N)]\in \mathbb{R}^{N+1}.
 \end{align}
 At training time the prediction of $x(1),...,x(N)$ is thus computed by convolving the input with the kernels in each layer $l=1,...,L$, followed by the $1\times 1$ convolution. At testing time a one-step ahead prediction $\hat x(t+1)$ for $(t+1)\geq r$ is given by inputting $[x(t+1-r),...,x(t)]$ in the trained model. An $n$-step ahead forecast is made sequentially by feeding each prediction back into the network at the next time step, e.g. a two-step ahead out-of-sample forecast $\hat x(t+2)$ is made using $[x(t+2-r),...,\hat x(t+1)]$.

The idea of the network is thus to use the capabilities of convolutional neural networks as autoregressive forecasting models. In a simple autoregressive model of order $p$ the forecasted value for $x(t+1)$ is given by $\hat x(t+1) = \sum_{i=1}^p \alpha_i x_{t-i} + \epsilon(t)$, where $\alpha_i$, $i=1,...,p$ are learnable weights and $\epsilon(t)$ is white noise. 
With the WaveNet model as defined above, the forecasted conditional expectation for every $t\in \{0,...,N\}$ is 
\begin{align}
\mathbb{E}[x(t+1)|x(t),...,x(t-r)] = \beta_1 (x(t-r)) + ... + \beta_r(x(t)),
\end{align}
 where the functions $\beta_i$, $i=1,...,r$ are data-dependent and optimized through the convolutional network. We remark that even though the weights depend on the underlying data, due to the convolutional structure of the network, the weights are shared across the outputted filter map resulting in a weight matrix that is translation-invariant. 

\paragraph{\textbf{Objective function.}}
The network weights, the filters $w^l_h$, are trained to minimize the mean absolute error (MAE); to avoid overfitting, i.e. too large weights, we use L2 regularization with regularization term $\gamma$, so that the cost function is given by
\begin{equation}\label{eq:obj}
E(w) = \frac{1}{N}\sum_{t=0}^{N-1}\left|\hat x(t+1) - x(t+1)\right| + \frac{\gamma}{2} \sum_{l=0}^{L}\sum_{h=1}^{M_{l+1}}(w^l_h)^2,
\end{equation}
where $\hat x(t+1)$ denotes the forecast of $x(t+1)$ using $x(0),...,x(t)$. Minimizing $E(w)$ results in a choice of weights that make a tradeoff between fitting the training data and being small. Too large weights often result in the network being overfitted on the training data, so the L2 regularization, by forcing the weights to not become too big, enables the model to generalize better on unseen data. 
\begin{remark}[Relation to the Bayesian framework] In a Bayesian framework minimizing this cost function is equivalent to maximizing the posterior distribution under a Laplace distributed likelihood function centered at the value outputted by the model $\hat x(t+1)$ with a fixed scale parameter $\beta = \frac{1}{2}$,
\begin{align}
p(x(t+1)|x(0),...,x(t),\theta)\sim Laplace(\hat x(t+1),\beta),
\end{align}
 and with a Gaussian prior on the model parameters.
 \end{remark}
  The output is obtained by running a forward pass through the network with the optimal weights being a point estimate from the posterior distribution. Since the MAE is a scale-dependent accuracy measure one should normalize the input data to make the error comparable for different time series. 

\paragraph{\textbf{Weight optimization.}}
The aim of training the model is to find the weights that minimize the cost function in \eqref{eq:obj}. A standard weight optimization is based on gradient descent in which one incrementally updates the weights based on the gradient of the error function,
\begin{align}\label{eq:weightupd}
w^l_h(\tau+1) = w^l_h(\tau) - \eta \nabla E(w(\tau)),
\end{align}
for $\tau = 1,...,T$, where $T$ is the number of training iterations and $\eta$ is the learning rate. Each iteration $\tau$ thus consists of a forward run in which one computes the forecasted vector $\hat x$ and the corresponding error $E(w(\tau))$, and a backward pass in which the gradient vector $\nabla E(w(\tau))$, the derivatives with respect to each weight $w^l_h$, is computed and the weights are updated according to \eqref{eq:weightupd}. The gradient vector is computed through backpropagation, which amounts to applying the chain rule iteratively from the error function computed in the final layer until the gradient with respect to the required layer weight $w^l_h$ is obtained:
\begin{align}
\frac{\partial E(w(\tau))}{\partial w^l_h(j,m)} = \sum_{i=1}^{N_l} \frac{\partial E(w(\tau))}{\partial f^{l}(i,h)}\frac{\partial f^l(i,h)}{\partial a^l(i,h)}\frac{\partial a^l(i,m)}{\partial w^l_h(j,m)},
\end{align}
 where we sum over all the nodes in which the weight of interest occurs. The number of training iterations $T$ is chosen so as to achieve convergence in the error. Here we employ a slightly modified weight update by using the Adam gradient descent \cite{kingma14}. This method computes adaptive learning rates for each parameter by keeping an exponentially decaying average of past gradients and squared gradients and use these to update the parameters. The adaptive learning rate allows the gradient descent to find the minimum more accurately.

\paragraph{\textbf{Activation functions.}} In each layer we use a non-linearity, or activation function, to transform the output from the convolution, hereby allowing the model to learn non-linear representations of the data. In our model the non-linearity takes the form of the rectified linear unit (ReLU) defined as $ReLU(x):=\max(x,0)$, so that the output from layer $l$ is:
\begin{equation}
f^l = \left[ ReLU(w^l_1 *_d f^{l-1}) + b,...,ReLU(w_{M_{l}}^l *_d f^{l-1} + b) \right],
\end{equation}
where $b\in\mathbb{R}$ denotes the bias that shifts the input to the nonlinearity, $*_d$ denotes as usual the convolution with dilation $d$ and $f^l\in \mathbb{R}^{1\times N_l\times M_{l+1}}$ denotes the output of the convolution with filters $w^h_l$, $h=1,...,M_{l}$ in layer $l$. Unlike the gated activation function used in  \cite{vanoord16} for audio generation, here we propose to use the ReLU as it was found to be most efficient when applied to the forecasting of the non-stationary, noisy time series. At the same time using the ReLU reduces the training time and thus simplifies the model. The final layer, $l=L$, has a linear activation function, which followed by the 1$\times$1 convolution then outputs the forecasted value of the time series $\hat x=[\hat x(0),...,\hat x(N)]$. 
 
When training a deep neural network, one of the problems keeping the network from learning the optimal weights is that of the vanishing/exploding gradient \cite{bengio94}\cite{glorot10}. As backpropagation computes the gradients by the chain rule, when the derivative of the activation function takes on either small or large values, multiplication of these numbers can result in the gradients for the weights in the initial layers to vanish or explode, respectively. This results in either the weights being updated too slowly due to the too small gradient, or not being able to converge to the minimum due to gradient descent step being too large. One solution to this problem is to initialize the weights of the convolutional layers in such a way that neither in the forward nor in the backward propagation of the network the weights reduce or magnify the magnitudes of the input signal and gradients, respectively. A proper initialization of the weights would keep the signal and gradients in a reasonable range of values throughout the layers so that no information will be lost while training the network. As derived in \cite{he15}, to ensure that the variance of the input is similar to the variance of the output, a sufficient condition is
\begin{align}
\frac{1}{2}z \mathrm{Var}[w^l_h]=1, \;\; \textnormal{for } h=1,...,M_{l+1}, \forall l,
\end{align}
which leads to a zero-mean Gaussian distribution whose standard deviation is $\sqrt{2/z}$, where $z$ is the total number of trainable parameters in the layer. In other words, the weights of the ReLU units are initialized (for $\tau=0$) as
\begin{equation}
w_h^l \sim \mathcal{N}\left(0,\sqrt{\frac{2}{z}}\right),
\end{equation}
with $z = M_{l}\cdot 1\cdot k$, the number of filters in layer $l$ times the filter size $1\times k$.

\paragraph{\textbf{Residual learning. }} When adding more layers to the network, standard backpropagation becomes unable to find the optimal weights, resulting in a higher \emph{training} error. This problem, called the degradation problem \cite{he15a}, is thus not caused by overfitting. Consider a shallow network with a small number of layers, and its deeper counterpart. The deeper model should not result in a higher training error, since there exists a solution by construction: set all the weights in the added layers to identity mappings. However in practice, gradient descent algorithms tend have problems learning the identity mappings. The proposed way around this problem is to use residual connections \cite{he15a} which force the network to approximate $\mathcal{H}(x)-x$, instead of $\mathcal{H}(x)$, the desired mapping, so that the identity mapping can be learned by driving all weights to zero. Optimizing the residual mapping by driving the weights to zero tends to be easier than learning the identity. The way residual connections are implemented is by using shortcut connections, which skip one or more layer(s) and thus get added unmodified to the output from the skipped layers. While in reality, the optimal weights are unlikely to be exactly the identity mappings, if the optimal function is closer to the identity than a zero mapping, the proposed residual connections will still aid the network in learning the better optimal weights. 

Similar to \cite{vanoord16}, in our network, we add a residual connection after each dilated convolution from the input to the convolution to the output. In the case of $M_l>1$ the output from the non-linearity is passed through a 1$\times$1 convolution prior to adding the residual connection. This is done to make sure that the residual connection and the output from the dilated convolution both have the same number of channels. This allows us to stack multiple layers, while retaining the ability of the network to correctly map dependencies learned in the initial layers.

\subsection{Relation to discrete wavelet transform}
The structure of the network is closely related to the discrete wavelet transform (DWT). Wavelet analysis can be used to understand how a given function changes from one period to the next by matching a wavelet function, of varying scales (widths) and positions, to the function. The DWT is a linear transform of $x=(x_t)_{t=0}^{N-1}$ with $N=2^J$ which decomposes the signal into its high- and low-frequency components by convolving it with high- and low-pass filters. In particular, at each level $j$ of the transform the input signal is decomposed into $N_j=\frac{N}{2^j}$ wavelet and scaling coefficients $\langle x, \psi_{j,k}\rangle$ and $\langle x, \phi_{j,k}\rangle$ (also called the approximation and detail) for $k=0,...,N_j-1$, by convolving the input $x$ simultaneously with filters $h$ and $g$ given by
\begin{align}
h(t) = 2^{-j/2}\psi(-2^{-j}t),\\
g(t) = 2^{-j/2}\phi(-2^{-j}t),\\
\end{align}
where $\psi(\cdot)$ is the wavelet and $\phi(\cdot)$ the scaling function. In every subsequent level we apply the transform to the approximation coefficients, in this way discarding the high-frequency components (the detail) and ending up with a smoothed version of the input signal. This is very similar to the structure of the CNN, where in each subsequent layer we convolve the input from the previous layer with a learnable filter. In each layer, the filter is used to recognize local dependencies in the data, which are subsequently combined to represent more global features, until in the final layer we compute the output of interest. By allowing the filter to be learnable as opposed to fixed \'a priori as is the case in the DWT, we aim to find the filter weights that minimize the objective function \eqref{eq:obj} by recognizing the certain patterns in the data in this way resulting in an accurate forecast of the time series.


\subsection{Conditioning}
When forecasting a time series $x=(x_t)_{t=0}^{N-1}$ conditional on another series $y=(y_t)_{t=0}^{N-1}$, we aim at maximizing the conditional likelihood,
\begin{equation}
p(x|y,\theta) = \prod_{t=0}^{N-1}p\left(x(t+1)|x(0),...,x(t),y(0),...,y(t),\theta\right).
\end{equation}
The conditioning on the time series $y$ is done by computing the activation function of the convolution with filters $w^1_h$ and $v^1_h$ in first layer as 
\begin{equation}
ReLU(w^1_h *_d x + b)+ReLU(v^1_h *_d y + b),
\end{equation}
for each of the filters $h=1,...,M_1$. When predicting $x(t+1)$ the receptive field of the network must contain only $x(0),...,x(t)$ and $y(0),...,y(t)$. Therefore, similar to the input, to preserve causality the condition is appended with a vector of zeros the size of the receptive field. 
In \cite{vanoord16} the authors propose to take $v^1_h$ as a $1\times 1$ filter. Given the short input window, this type of conditioning is not always able to capture all dependencies between the time series. Therefore, we use a $1\times k$ convolution, increasing the probability of the correct dependencies being learned with fewer layers. The receptive field of the network thus contains $2^{L-1}k$ elements of both the input and the condition(s). 

\begin{figure}[h!]
\centering
\includegraphics[width=0.9 \textwidth]{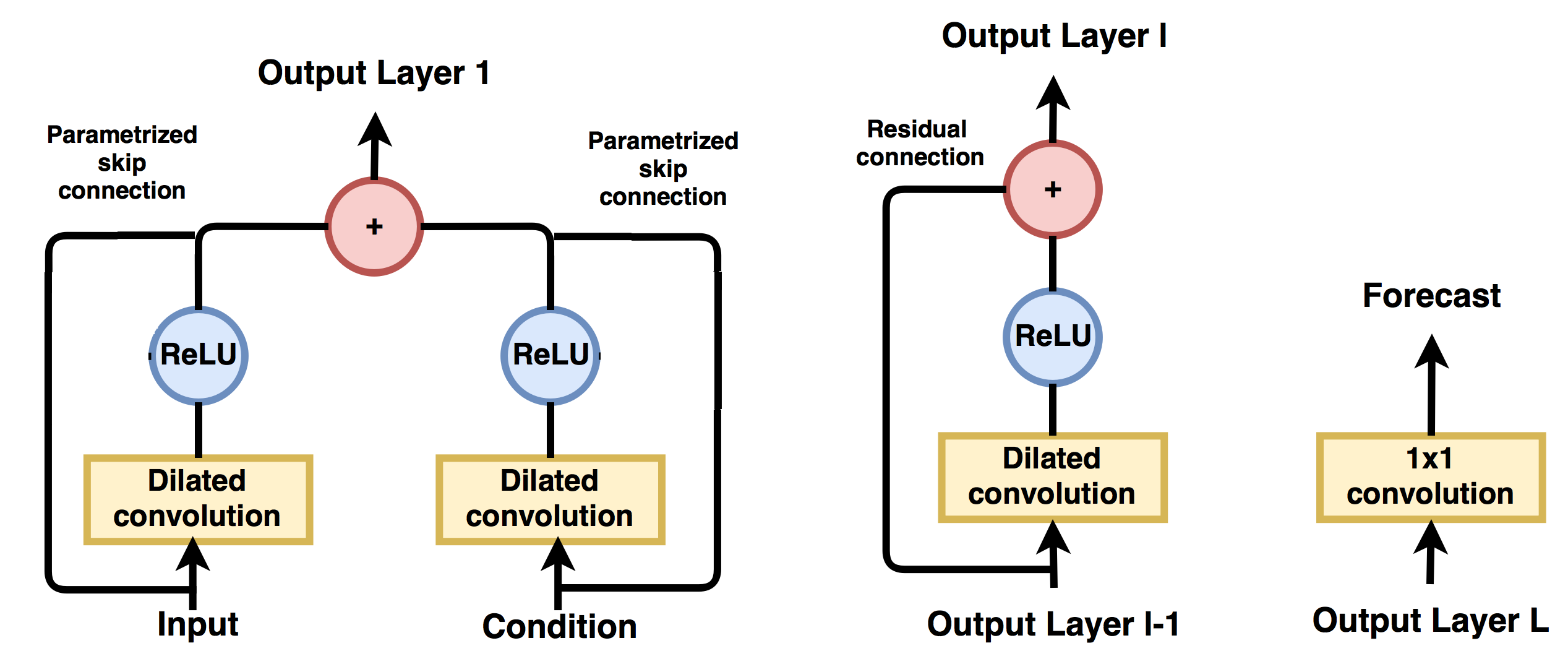}\hspace{0.5cm}     
\caption{The network structure. In the first layer (L) the input and condition (with the zero padding) are convolved, passed through the non-linearity and summed with the parametrized skip connections. The result from this first layer is the input in the subsequent dilated convolution layer with a residual connection from the input to the output of the convolution. This is repeated for the other layers, until we obtain the output from layer $L$ (M). This output is passed through a 1$\times$1 convolution, resulting in the final output: the forecasted time series (R).}\label{figtotalnet}
\end{figure}

Instead of the residual connection in the first layer, we add skip connections parametrized by 1$\times$1 convolutions from both the input as well as the condition to the result of the dilated convolution. The conditioning can easily be extended to a multivariate $M \times N$ time series by using $M$ dilated convolutions from each separate condition and adding them to the convolution with the input. The parametrization of the skip connections makes sure that our model is able to correctly extract the necessary relations between the forecast and both the input and condition(s). Specifically, if a particular condition does not improve the forecast, the model can simply learn to discard this condition by setting the weights in the parametrized skip connection (i.e. in the 1$\times$1 convolution) to zero.  This enables the conditioning to boost predictions in a discriminative way. If the number of filters $M_l$ is larger than one, the parametrized skip connection uses a 1$\times$1 convolution with $M_l$ filers, so that the summation of the skip connection and the original convolution is valid. The network structure is shown in Figure \ref{figtotalnet}. 

\begin{remark}[Ability to learn non-linear dependencies]\label{rem1} We remark here on the ability of the model to learn non-linear dependencies in and between time series. A feedforward neural network requires at least a single hidden layer with a sufficiently large number of hidden units in order to approximate a non-linear function \cite{hornik91}. If in the CNN we set the filter width to one, a necessary requirement for the model to learn non-linear dependencies will be to have $M_l>1$, since in this case the role of the filters is similar to that of the hidden units. Alternatively, learning non-linearities in a CNN requires the use of both a filter width and number of layers larger than one. Each layer essentially computes a dot-product and a summation of a non-linear transformation of several outputs in the previous layer. This output is in turn a combination of the input and condition(s) and the role of the hidden units is played by the summation over the filter width, hereby allowing non-linear relations to be learned in and between the time series.
\end{remark}

\section{Experiments}
Here, we evaluate the performance of the proposed Wavenet architecture versus current state-of-the-art models (RNNs and autoregressive models) when applied to learning dependencies in chaotic, non-linear time series. The parameters in the model, unless otherwise mentioned, are set to $k=2$, $L=4$, $M_l=1$ for $l=0,...,L-1$, the Adam learning rate is set to 0.001 and the number of training iterations is 20000. The regularization rate is chosen to be 0.001. We train networks with different random seeds, discard any network which underperforms already on the training set and report the average results on the test set over three selected networks. 

\subsection{An artificial example}\label{sec41}
In order to show the ability of the model to learn both linear and non-linear dependencies in and between time series, we train and test the model on the chaotic Lorenz system. The Lorenz map is defined as the solution $(X,Y,Z)$ to a system of ordinary differential equations (ODEs) given by
\begin{align}
&\dot X = \sigma(Y-X)\\
&\dot Y = X(\rho-Z)-Y\\
&\dot Z = XY - \beta Y,
\end{align}
with initial values $(X_0,Y_0,Z_0)$. We approximate the solution using an Euler method. We present in Table \ref{tab1} the one-step ahead forecasting results for each of the three coordinates $(X,Y,Z)$ with the unconditional WaveNet (uWN) and the conditional Wavenet (cWN). In the cWN the forecast of e.g. $\hat X_t$ contains $X_{t-1},...,X_{t-1-r}$, $Y_{t-1},...,Y_{t-1-r}$ and $Z_{t-1},...,Z_{t-1-r}$. We use a training time series of length 1000, i.e. $(X_t)_{t=1}^{1000}$, $(Y_t)_{t=1}^{1000}$ and $(Z_t)_{t=1}^{1000}$. Then we perform a one-step ahead forecast of $X_t$, $Y_t$ and $Z_t$ for $t=1000,...,1500$, and compare the forecasted series $\hat X_t$, $\hat Y_t$ and $\hat Z_t$ to the true series. The RMSE is computed over this test set. Comparing the results of the uWN with the RMSE benchmark of $0.00675$ from \cite{hsu17} obtained with an Augmented LSTM, we conclude that the network is well-capable of extracting both linear and non-linear relationships in and between time series. At the same time, conditioning on other related time series reduces the standard deviation as one can see from the smaller standard deviation in the RMSE of the cWN compared to the uWN. 

\begin{table}[h!]
\begin{center}\label{tab1}
\begin{tabular}{c|c|c}
Coordinate & RMSE uWN & RMSE cWN\\ \hline\hline
X & 0.00577 (0.00242) & 0.00174 (0.00133) \\
Y & 0.00864 (0.00487) & 0.00583 (0.00350)\\
Z & 0.00496 (0.00363) & 0.00536 (0.00158)\\
\end{tabular}
\caption{RMSE (mean (standard deviation)) for the one-step ahead forecast of the Lorenz map with $(X_0,Y_0,Z_0) = (0,1,1.05)$, $\sigma = 10$, $\rho - 28$ and $\beta = 8/3$. The cWN results for each coordinate are conditioned on the other two in the system. The current benchmark, the average RMSE over X, Y and Z, is $0.00675$ from \cite{hsu17}.}
\end{center}
\end{table}

\begin{figure}[h!]
\centering
\includegraphics[width=0.5 \textwidth]{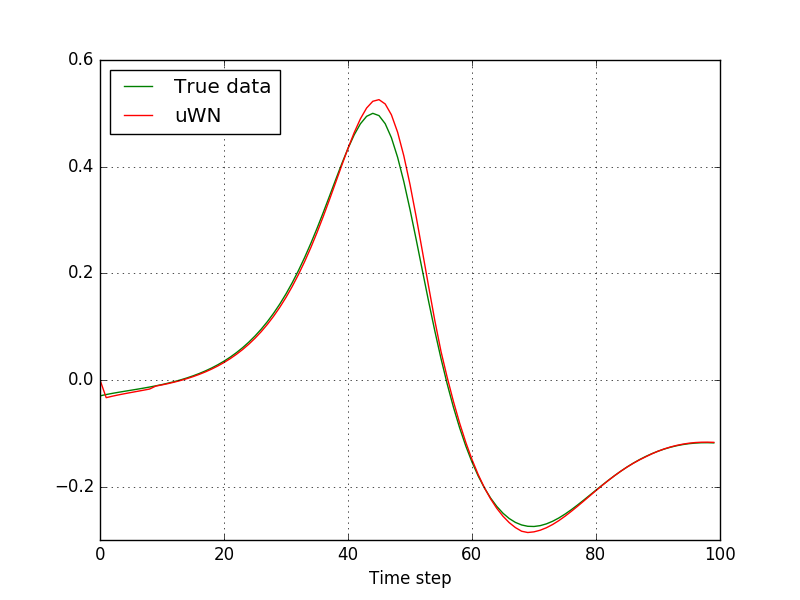}\hspace{-0.5cm}     
\includegraphics[width=0.5 \textwidth]{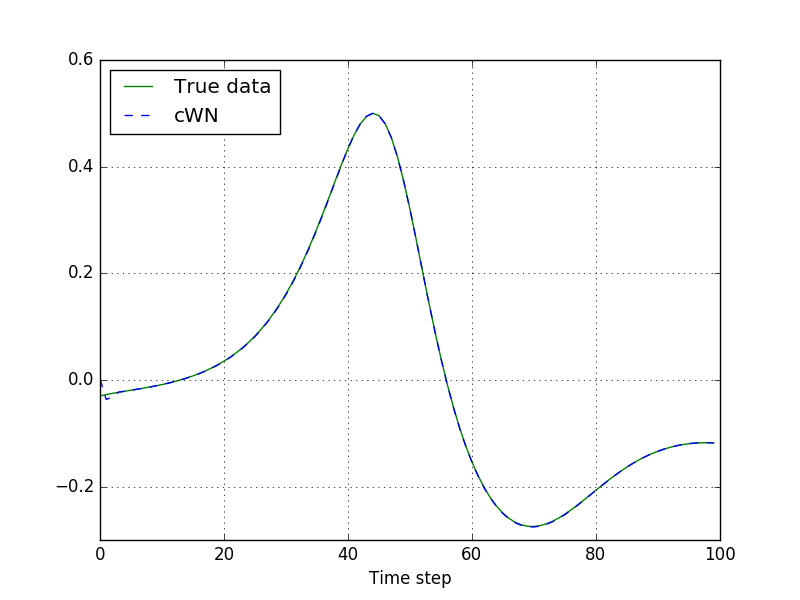}\hspace{-0.5cm}   
\includegraphics[width=0.5 \textwidth]{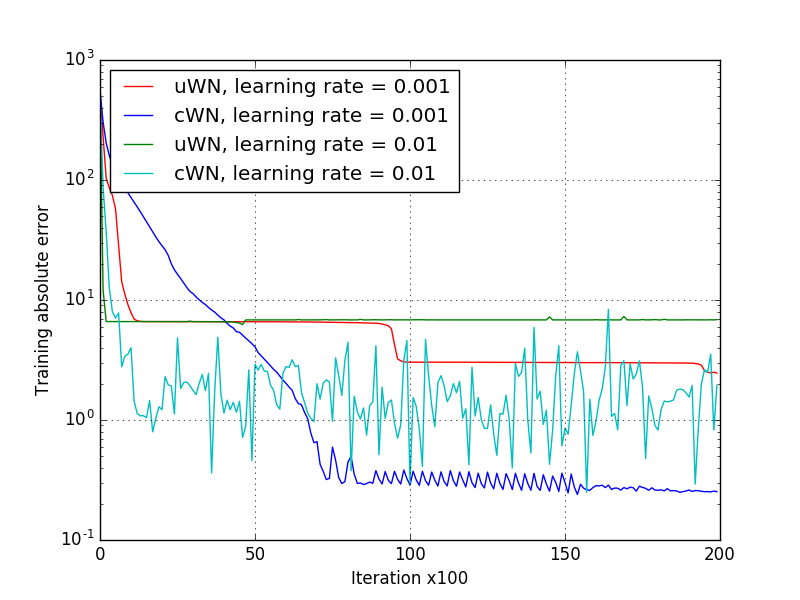}\hspace{-0.5cm}
\includegraphics[width=0.5 \textwidth]{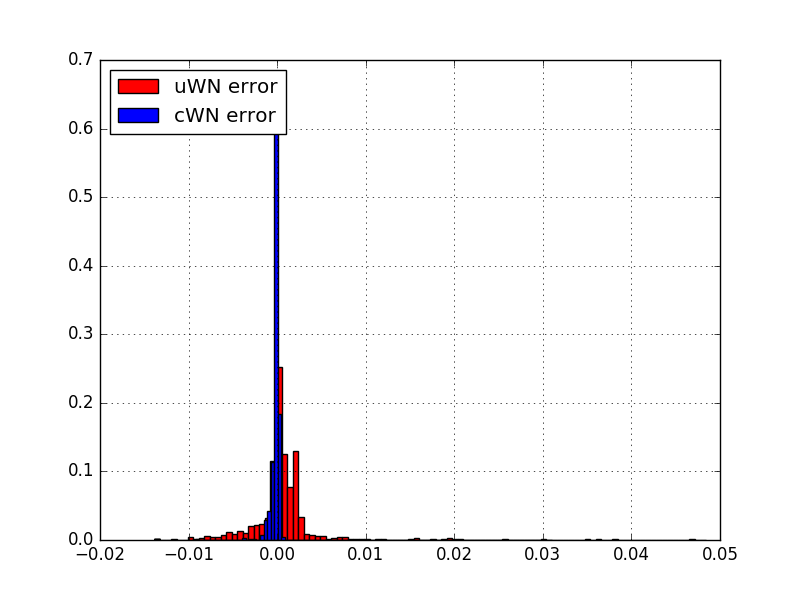}\hspace{-0.5cm}    
\caption{The $X$-coordinate of the Lorenz map (green), the unconditional one-step ahead forecast (red) (TL), the conditional forecast (blue) (TR), the convergence behaviour of unconditional and conditional forecast for different learning rates (LL) and the histogram of the errors for the one-step-ahead forecast on the test set (LR).}\label{fig1}
\end{figure}

\begin{figure}[h!]
\centering
\includegraphics[width = 0.5 \textwidth]{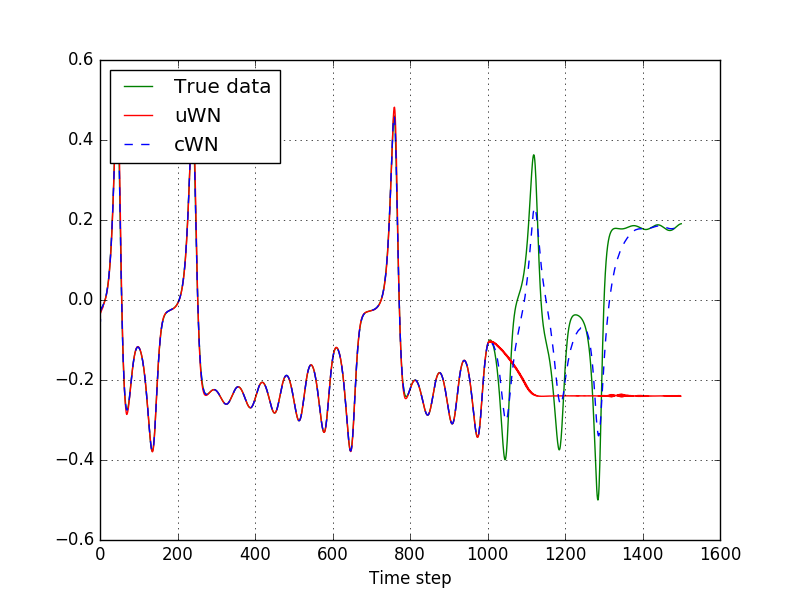}\hspace{-0.5cm}
\includegraphics[width = 0.5 \textwidth]{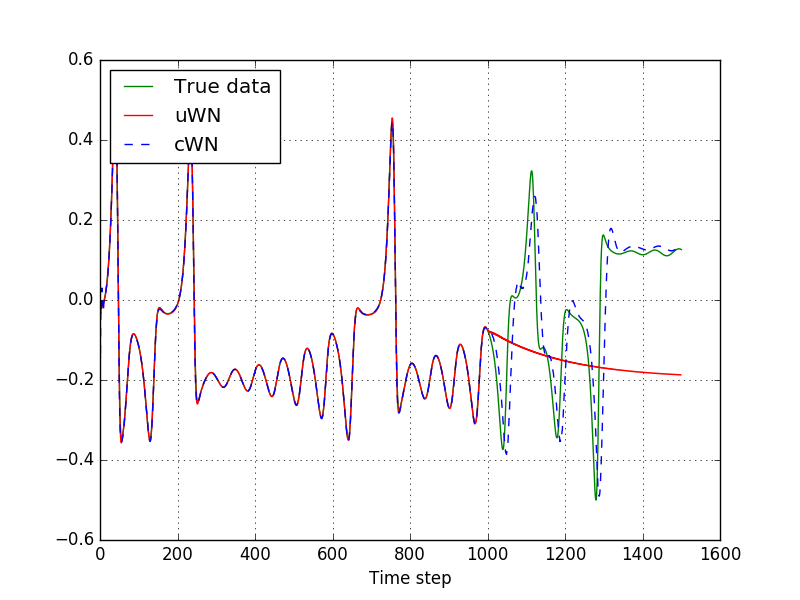}\hspace{-0.5cm}
\caption{The training sample $t\in [0,1000]$ and a fully out-of-sample forecast for time steps $t\in[1000, 1500]$ for the $X$-coordinate (L) and the $Y$-coordinate (R)}\label{fig2}
\end{figure}

\begin{figure}[h!]
\centering
\includegraphics[width = 0.5 \textwidth]{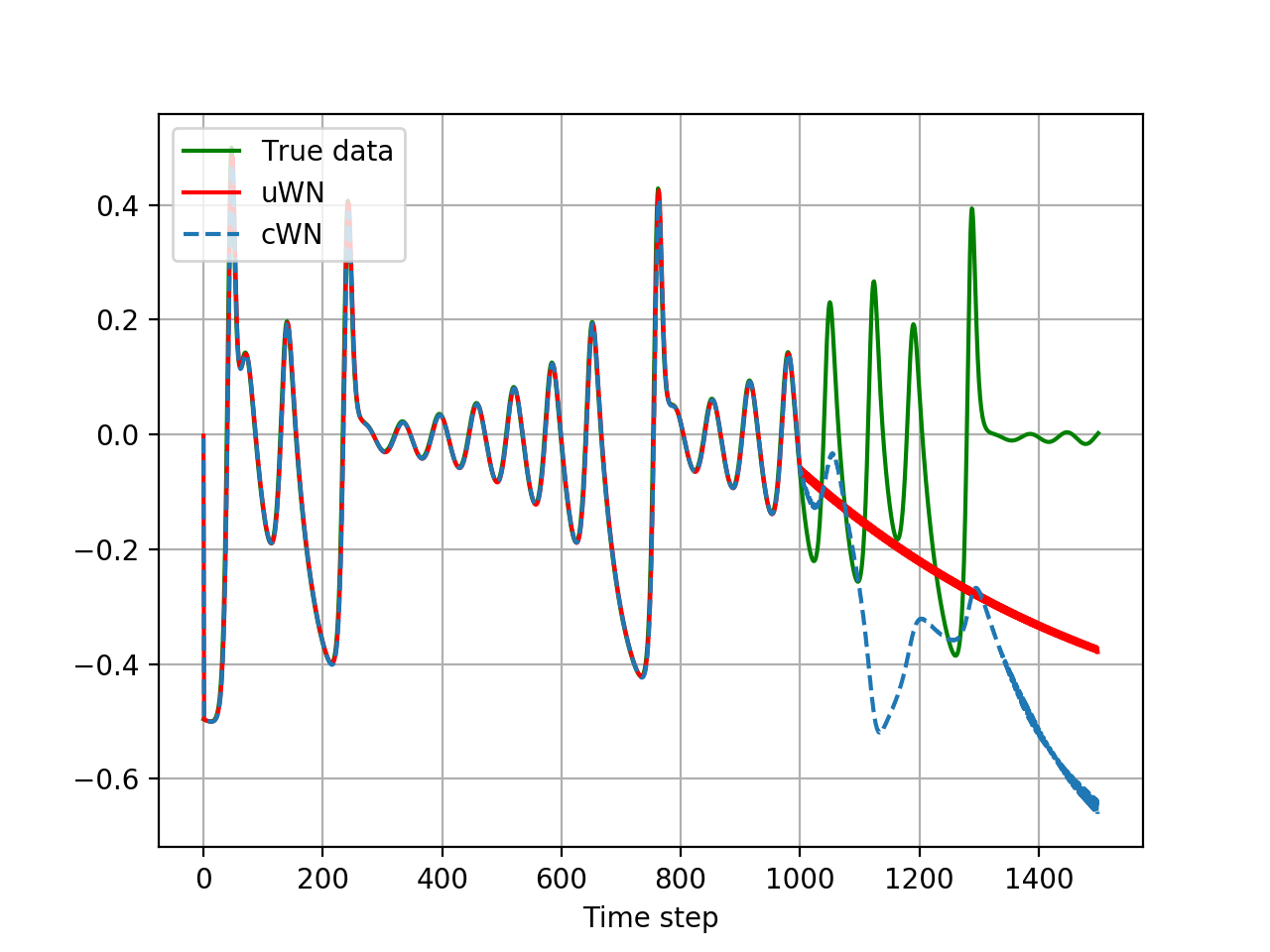}\hspace{-0.5cm}
\includegraphics[width = 0.5 \textwidth]{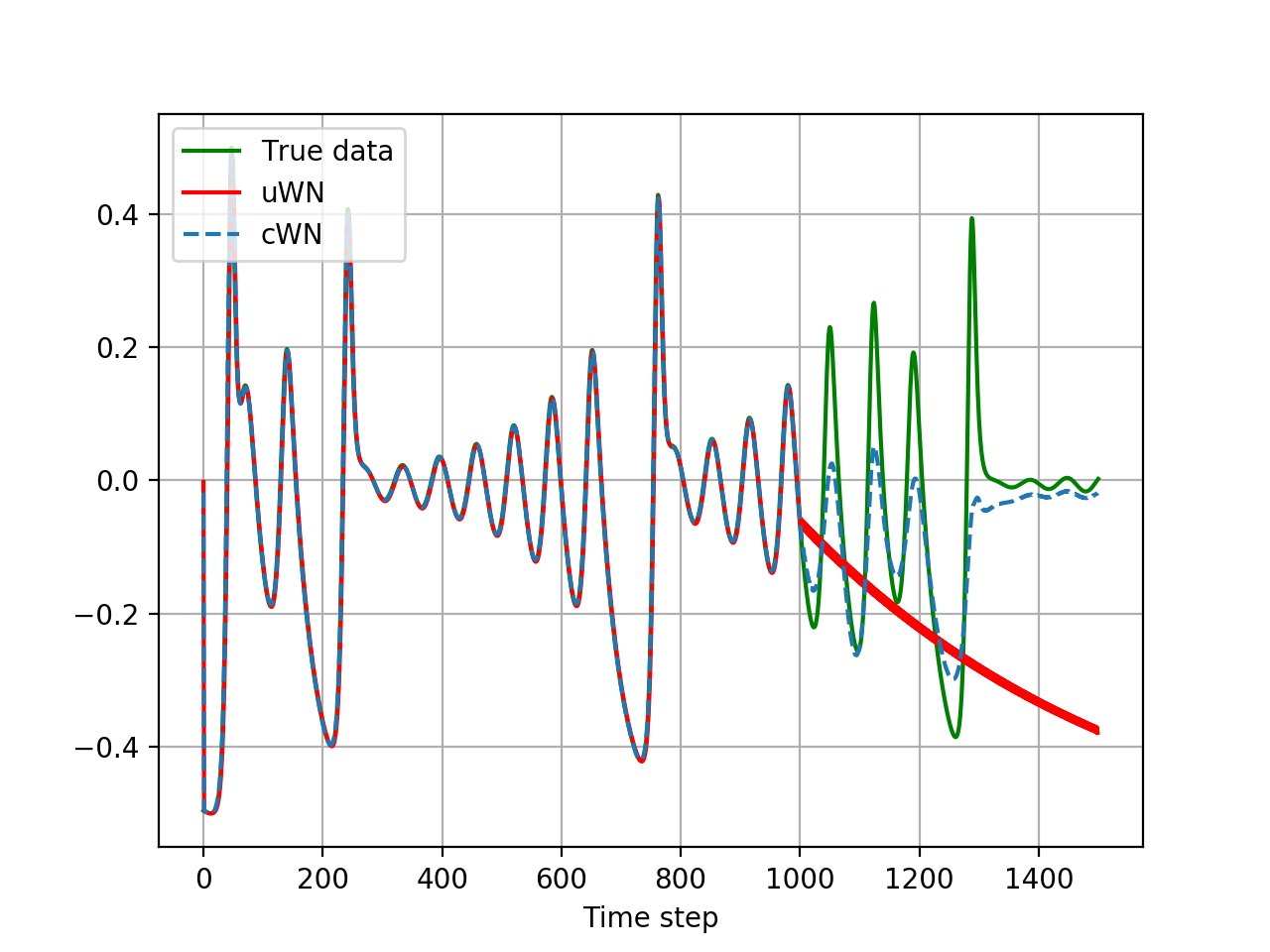}\hspace{-0.5cm}
\includegraphics[width = 0.5 \textwidth]{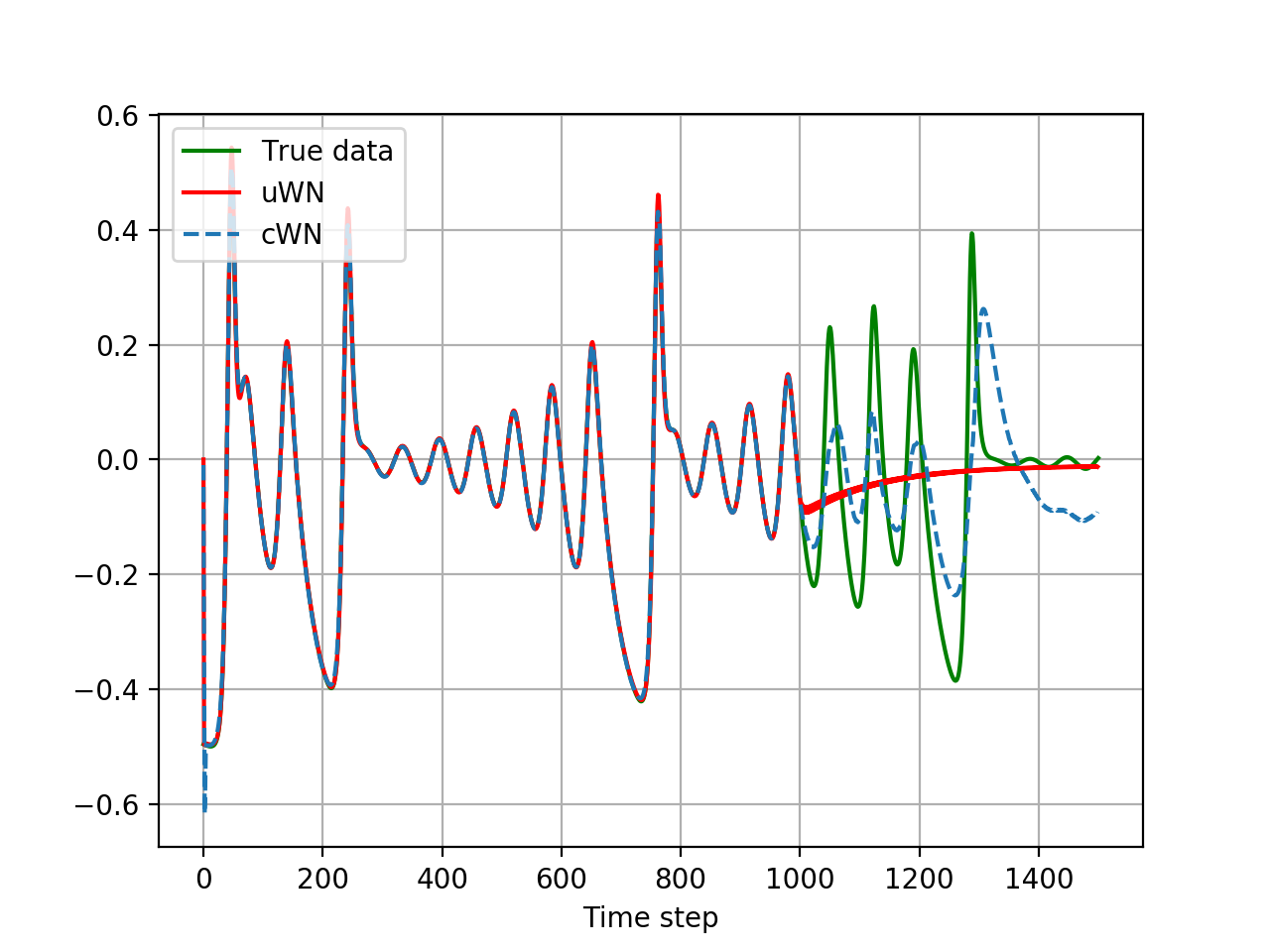}\hspace{-0.5cm}
\includegraphics[width = 0.5 \textwidth]{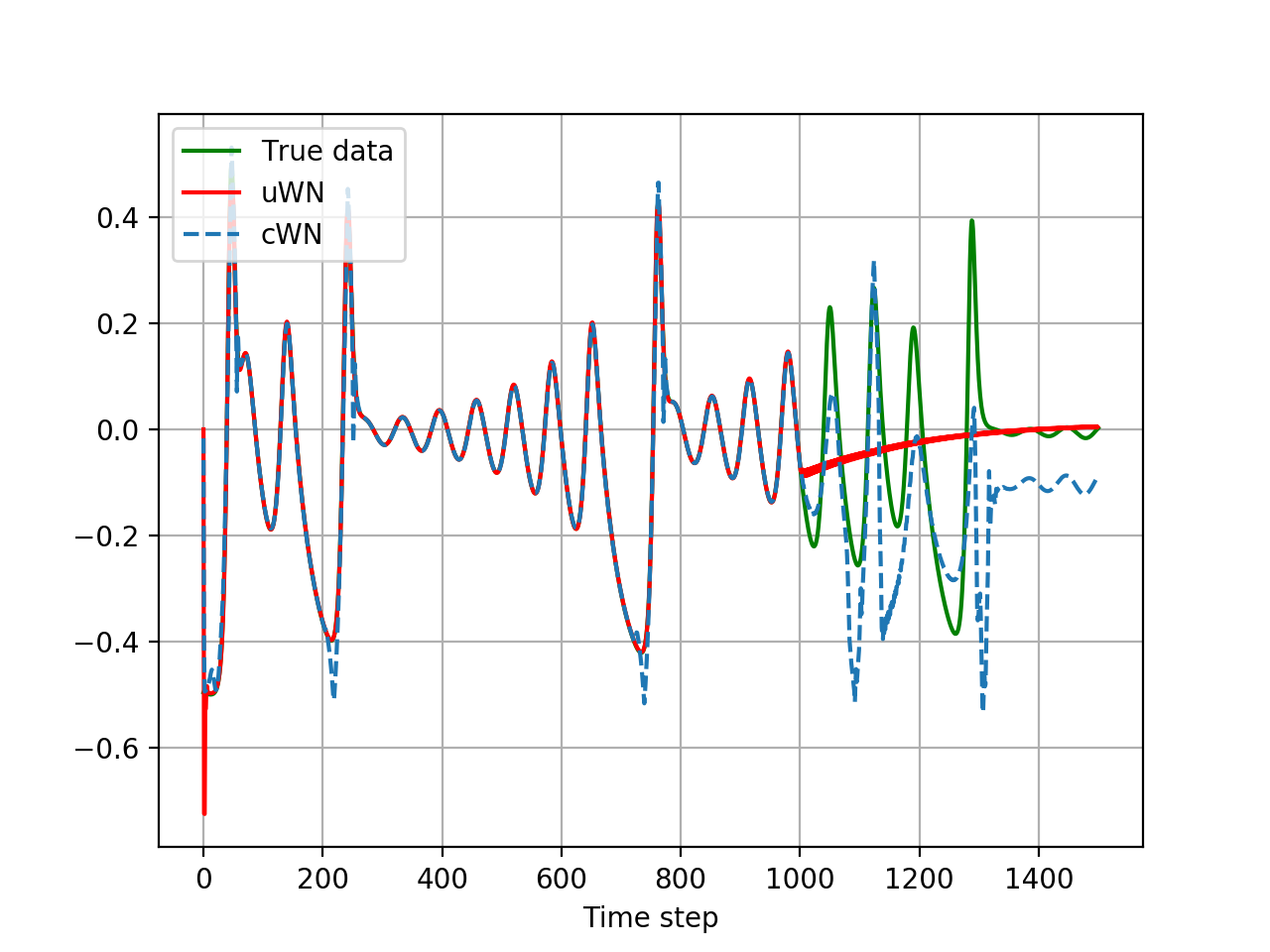}\hspace{-0.5cm}
\caption{The training sample $t\in [0,1000]$ and a fully out-of-sample forecast for time steps $t\in[1000, 1500]$ for the $Z$-coordinate with $M_l=1$, $k=1$ (TL), $M_l=1$, $k=2$ (TR), $M_l=3$, $k=1$ (LL) and $M_l=3$, $k=2$ (LR).}\label{fig2a}
\end{figure}

In Figure \ref{fig1} we show the forecast of the $X$-coordinate in more detail. As seen from both the forecast and the histogram of the error, the cWN results in a more precise forecast. Furthermore, the learning rate of 0.001, while resulting in a slower initial convergence, is much more effective at obtaining the minimum training error, both unconditionally as well as conditionally. Figure \ref{fig2} shows the out-of-sample forecast of the uWN and the cWN for X and Y. Conditioning allows the network to learn the true underlying dynamics of the system, resulting in a much better out-of-sample forecast. From the RMSE in Table \ref{tab1} we see that the conditioning does not improve the accuracy of one-step-ahead forecast in the case of $Z$ (as the forecast might already unconditionally be very accurate), however from the out-of-sample forecast plots in Figure \ref{fig2} and the top right plot in Figure \ref{fig2a} we can conclude that conditioning is necessary in order to learn the underlying non-linear and linear dynamics in between the series. Furthermore, with Figure \ref{fig2a} we verify Remark \ref{rem1} numerically. Using only one filter and a filter width of one does not allow the non-linear dependencies to be learned correctly, while using either a filter width or the number of filters larger than one significantly improves the out-of-sample forecast. Unfortunately, using both $k=2$ and $M_l=3$ results in a worsened performance of the forecast, since the combination of the wide receptive field and a large number of parameters results in the model being unable to find the optimal weights. 

\subsection{Financial data}
We analyze the performance of the network on the S\&P500 data in combination with the volatility index and the CBOE 10 year interest rate to analyze the ability of the model to extract -- both unconditionally as well as conditionally-- meaningful trends and relationships in and between the noisy datasets. Furthermore, we test the performance on several exchange rates to showcase the ability of the model to efficiently learn long-term dependencies.
\subsubsection{Data preparation}
We define a training period of 750 days (approximately three years) and a testing period of 250 days (approximately one year) on which we perform the one-day ahead forecasting. The data from 01-01-2005 until 31-12-2016 is split into nine of these periods with non-overlapping testing periods. Let $P^s_t$ be the value of time series $s$ at time $t$. We define the return for $s$ at time $t$ over a one-day period as
\begin{align}
R_t^s = \frac{P_t^s-P_{t-1}^s}{P^s_{t-1}}.
\end{align} 
Then we normalize the returns by subtracting the mean, $\mu_{train}$, and dividing by the standard deviation, $\sigma_{train}$, obtained over all the time series that we will condition on in the training period (note that using the mean and standard deviation over the train and test set would result in look-ahead biases). The normalized return is then given by
\begin{align}
\tilde R_t^s = \frac{R_t^s - \mu_{train} }{\sigma_{train}}.
\end{align}
We then divide the testing periods into three main study periods: period A from 2008 until 2010, period B from 2011 until 2013 and period C from 2014 until 2016. The performance is then evaluated by performing one-step ahead forecasts over these testing periods and comparing the MASE scaled by a naive forecast and the HITS rate. An MASE smaller than one means that the absolute size of the forecasted return is more accurate than that of a naive forecast, while a high HITS rate shows that the model is able to correctly forecast the direction of the returns. 

\subsubsection{Benchmark models}
We compare the performance of the WaveNet model with several well-known benchmarks: an autoregressive model widely used by econometricians, and an LSTM \cite{hochreiter97}\cite{fisher17}, currently the state-of-the-art in time series forecasting. Similar to \cite{fisher17} the LSTM is implemented using one LSTM layer with 25 hidden neurons and a dropout of 0.1 followed by a fully connected output layer with one neuron and we use 500 training epochs. LSTM networks require sequences of input features for training, and we construct the sequences using $r=2^{L-1}k$ historical time steps so that the receptive field of the WaveNet model is the same as the distance that the LSTM can see into the past. The LSTM is implemented to take as input a matrix consisting of sequences of all the features (the input and condition(s)), so that its performance can be compared to that of the VAR and the cWN. 

\subsubsection{Results} 
\paragraph{\textbf{Index forecasting}}
We compare the performance of the unconditional and the conditional WaveNet in forecasting the S\&P500, in the cWN case conditioned on both the volatility index and the CBOE 10 year interest rate. Using one filter and multiple layers should enable the model to learn non-linear trends and dependencies in and between the time series, and in this example we try to verify this numerically. From Table \ref{finalcomp} we see that the unconditional WaveNet performs best in terms of MASE. The conditional WaveNet exploits the correlation between the three time series resulting in a higher hit rate, but a slightly worse MASE compared to the unconditional one as it is fitted on multiple noisy series. The LSTM also performs similar to the cWN in terms of the HITS rate, but results in a higher MASE, meaning that both networks are able to forecast the direction of the returns, but the LSTM is worse at predicting the size of the return. The WaveNet model outperforms the VAR conditionally in period A terms of HITS rate, showcasing the ability of the model to learn relationships that are more complex than simple linear dependencies, \emph{if} these are present. After 2010 the dependencies between the S\&P500 and the interest rate and volatility index seem to have weakened (due to e.g. the lower interest rate or higher spreads) as the improvement of the conditional WaveNet over the unconditional WaveNet is smaller. This suggests that the WaveNet can be used to recognize these switches in the underlying financial regimes. Overall, in terms of the HITS rate the WaveNet performs similar to the state-of-the-art LSTM, in particular in period A, when strong dependencies were still present between the index, interest rate and volatility. In the other two periods the performance of the cWN in terms of the HITS is similar to that of a naive and the autoregressive forecast, from which we infer that there are no longer strong dependencies present between the time series. Furthermore, the good performance of the naive model in periods B and C can be explained by the fact that it implicitly uses the knowledge that the period after the financial crisis was a bull market with a rising price trend. From these results we can conclude that the WaveNet is indeed able to recognize patterns in the underlying datasets, if these are present. If not, the WaveNet model does not overfit on the noise in the series, as can be seen by the consistently lower MASE compared to the other models. 

\begin{table}[h!]
\begin{center}
\begin{tabular}{c|c|c|c|c|c|c|c|c|c|c}
 & \multicolumn{2}{|c|}{A} & \multicolumn{2}{|c|}{B} & \multicolumn{2}{|c|}{C}\\\hline\hline
Model & MASE & HITS & MASE & HITS & MASE & HITS\\\hline
Naive & 1 & 0.513&1 & \textbf{0.504}&1 & \textbf{0.555}\\
VAR & \textbf{0.698} & 0.507 & 0.701 & \textbf{0.505}  & 0.696 & 0.551 \\
LSTM &  0.873(0.026)&\textbf{0.525(0.006)} & 1.067(0.021)&0.496(0.016) & 0.929(0.021)&0.531(0.008) \\
uWN & \textbf{0.685(0.025)}& 0.515(0.007) & \textbf{0.681(0.002)}& 0.484(0.007) & \textbf{0.684(0.006)}&0.537(0.011)\\
cWN & 0.699(0.042) &\textbf{0.524(0.009)} & 0.693(0.014) & \textbf{0.500(0.009)}& 0.701(0.015) & 0.536(0.016)\\
\end{tabular}
\caption{MASE and HITS (mean(standard deviation)) for a one-step ahead forecast over the periods A, B and C of the S\&P500, both unconditional and conditional on the volatility index and the CBOE 10 year interest rate.}\label{finalcomp}
\end{center}
\end{table}

\begin{figure}[h!]
\centering
\includegraphics[width=0.32 \textwidth]{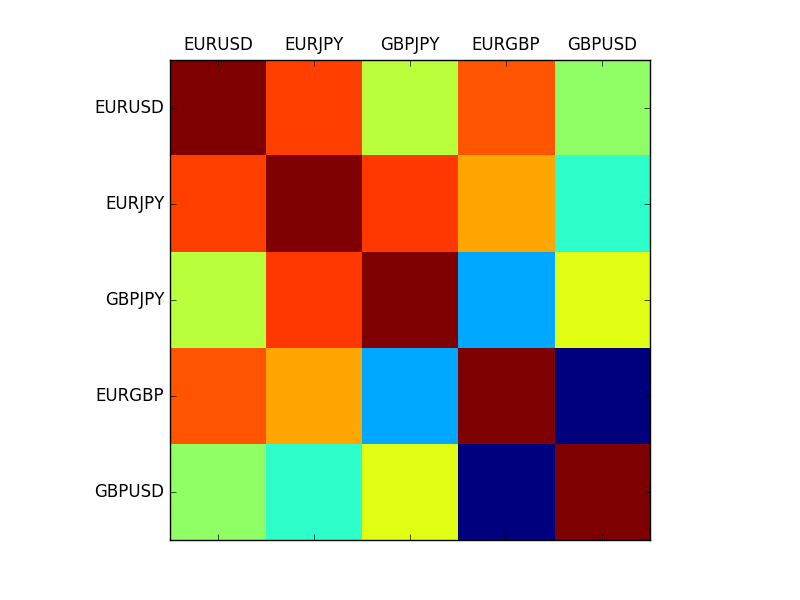}\hspace{-1cm}  
\includegraphics[width=0.32 \textwidth]{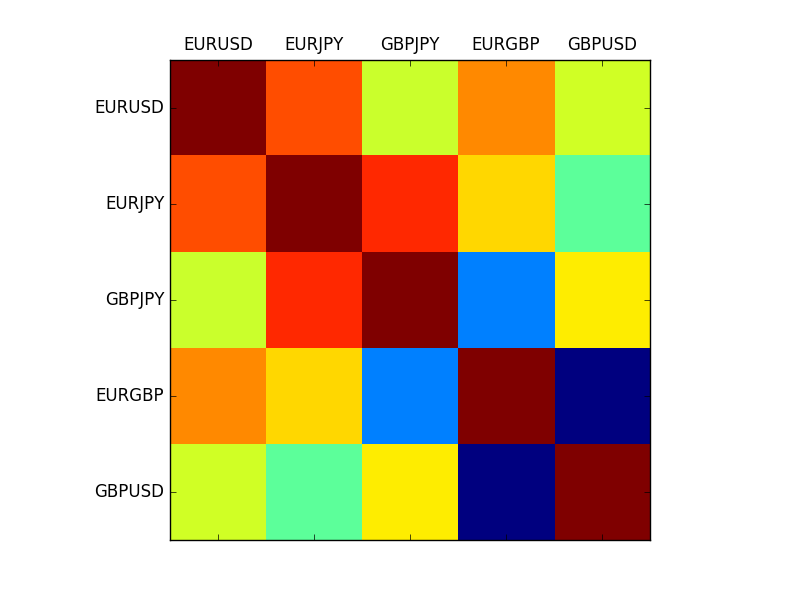}  \hspace{-1cm}  
\includegraphics[width=0.32 \textwidth]{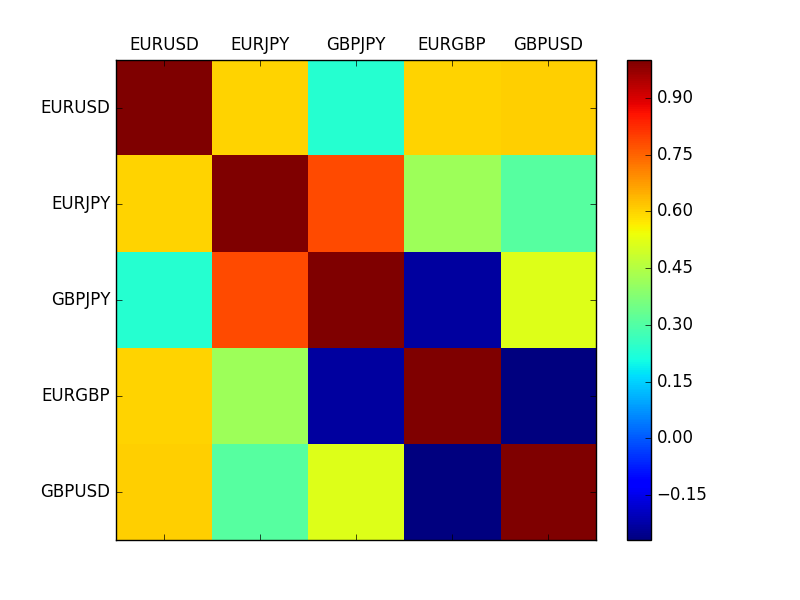}   
\caption{The correlation matrices of the five foreign exchange rates in periods A, B and C.}\label{corrmat}
\end{figure}

\paragraph{\textbf{Exchange rate data}}
Next we analyze the performance of the cWN on several exchange rates, in particular to compare the ability to discriminate between multiple inputs and  the ability to learn long-term dependencies of the proposed model versus the VAR and the LSTM. We present a statistical analysis of the exchange rates in Table \ref{tabstat}. Of particular relevance to the performance of the model are the standard deviation, skewness and kurtosis. A high standard deviation means that there is a lot of variance in the data. This could cause models to underperform as they become unable to accurately forecast the rapid movements. A high positive or negative skewness, meaning the asymmetry of the returns around its mean value, indicates the existence of a long right or left tail, respectively. We train the neural network to fit a symmetric distribution centered at the mean of the dataset. The existence of this tail could result in the trained model performing worse in cases of high absolute skewness. Kurtosis is a measure of the tails of the dataset compared to those of a normal distribution. A high kurtosis is the result of infrequent extreme deviations. If a model tends to overfit the dataset, and in particular overfit on these extreme deviations, a high kurtosis would result in a worse performance. Figure \ref{corrmat} shows the correlations between the exchange rates in the three periods. As expected, the exchange rates that contain the same currencies exhibit stronger correlations than those with different currencies. 

\begin{table}[h!]
\begin{center}
\begin{tabular}{c|c|c|c|c|c|c|c|c|c|c|c|c}
 Stock & \multicolumn{3}{|c|}{Mean Return} & \multicolumn{3}{|c|}{Standard deviation} & \multicolumn{3}{|c|}{Skewness} & \multicolumn{3}{|c|}{Kurtosis}\\\hline\hline
&A&B&C&A&B&C&A&B&C&A&B&C\\\hline
EURUSD & -0.022 & 0.017&-0.061&1.751&0.572&0.942&0.864&0.165&0.090&25.91&1.706&1.910\\
EURJPY & -0.050&0.053&-0.049&2.867&0.806&1.049&1.448&0.080&-0.685&43.96&1.246&5.556\\
GBPJPY & -0.110&0.048&-0.044&2.067&0.686&1.209&-0.009&0.387&-1.129&17.23&2.743&13.45\\
EURGBP & 0.045&0.018&-0.022&1.623&0.436&0.915&1.092&-0.229&0.444&26.60&0.606&4.919 \\
GPBUSD & -0.073&0.012&-0.058&0.975&0.453&0.895&-0.257&0.039&-1.289&1.616&0.591&15.06
\end{tabular}
\caption{Statistical analysis of five foreign exchange rates.}\label{tabstat}
\end{center}
\end{table}

\begin{table}[h!]
\begin{center}
\begin{tabular}{c|c||c|c|c|c|c}
Model & Period & EURUSD & EURJPY & GBPJPY & EURGBP & GBPUSD\\\hline\hline
VAR & A & 1.105 & 1.176 & 1.446 & 1.348 & 1.832\\
& B & 0.758 & 0.782 & 0.756 & 0.768 & 0.731 \\
& C & 0.716 & 0.738 & 0.737 &\textbf{ 0.709} & \textbf{0.713} \\\hline
LSTM & A & 0.829(0.012) & 0.863(0.005) & 0.880(0.004) & 0.868(0.005)&\textbf{0.893(0.007)}\\
& B & 0.925(0.024)& 0.911(0.029)&0.974(0.029)&0.948(0.023)&0.934(0.014)\\
& C & 0.950(0.016)&1.031(0.022) & 0.980(0.034)&0.839(0.034)&0.898(0.017)\\\hline
cWN &A& \textbf{0.693(0.016)} &\textbf{0.667(0.021)}&\textbf{0.759(0.064)}&\textbf{0.728(0.014)} &\textbf{0.834(0.089)}\\
& B &\textbf{0.690(0.006)} &\textbf{0.693(0.006)} &\textbf{0.699(0.005)} &\textbf{0.717(0.015)} &\textbf{0.710(0.009)} \\
& C& \textbf{0.702(0.009)}&\textbf{0.716(0.029)}&\textbf{0.721(0.014)}&\textbf{0.709(0.004)}&\textbf{0.716(0.004)}
\end{tabular}
\caption{MASE (mean(standard deviation)) one-step ahead multivariate forecast over the periods A, B and C of five foreign exhange rates.}\label{tabres}
\end{center}
\end{table}

In Table \ref{tabres} we present the results of the conditional WaveNet forecast over the exchange rate data, conditioning on the other exchange rates. Exchange rate data tends to contain long-term dependencies, so we expect the WaveNet model, with its ability of learning long term relationships, to perform well. As we see from the table, the WaveNet consistently outperforms the vector-autoregressive model and the LSTM in terms of the MASE. In period A the data has a very high kurtosis, probably due to the global financial crisis that was happening in 2008. Remarkably, we note that while the autoregressive model during this period of very high kurtosis performs worse than a naive forecast, the WaveNet model does not overfit the extremes resulting in a good performance in terms of the MASE. In periods of high absolute skewness and high standard deviation, but relatively low kurtosis, e.g. period C, the WaveNet model and the autoregressive model seem to be performing more or less equally. In period B we observe a relatively low standard deviation, low kurtosis and a small absolute skewness. In this period the WaveNet model is better able to extract the underlying dynamics compared to both the autoregressive model and the LSTM. We conclude that the WaveNet model is indeed able to extract long-term relationships, if present. In periods of high kurtosis it is still able to generalize well, while when the data has a high standard deviation and a high absolute skewness, i.e. in situations with many outliers, the model is unable to correctly forecast these outliers, causing the performance to be similar to that of a linear autoregressive model. Furthermore, as we see from Figure \ref{corrmat} some pairs of exchange rates have lower correlations than others. While the autoregressive model, when having as input both correlated as well as uncorrelated time series, tends to overfit, the WaveNet is better able to discriminate between the conditions by simply discarding those that do not improve the forecast, as can be seen by the consistently lower MASE. 

\section{Discussion and conclusion}
In this paper we presented and analysed the performance of a method for conditional time series forecasting based on a convolutional neural network known as the WaveNet architecture \cite{vanoord16}. The network makes use layers of dilated convolutions applied to the input and multiple conditions, in this way learning the trends and relations in and between the data. We analysed the performance of the WaveNet model on various time series, and compared the performance with the current state-of-the-art method in time series forecasting, the LSTM model, and a linear autoregressive model. We conclude that even though time series forecasting remains a complex task and finding one model that fits all is hard, we have shown that the WaveNet is a simple, efficient and easily interpretable network that can act as a strong baseline for forecasting. Nevertheless there is still room for improvement. One way of improving the ability of the CNN to learn non-linear dependencies is to use a large number of layers and filters. As we saw from Figure \ref{fig2a} one encounters the problem of a trade-off between the ability to learn non-linearities, which requires a large number of layers and filters, and that of overfitting, since a large number of layers results in a large receptive field and many parameters. This problem of the imbalance between the need of memory and the non-linearities was also adressed in \cite{binkowski17} by using a combination of an autoregressive model and a CNN. An alternative solution to this problem might be to use the parametrized skip connections in combination with an adaptive filter and will be studied in our further work. Furthermore, the WaveNet model proved to be a strong competitor to LSTM models, in particular when taking into consideration the training time. While on the relatively short time series the prediction time is negligible when compared to the training time, for longer time series the prediction of the autoregressive model may be sped up by implementing a recent variation that exploits the memorization structure of the network, see \cite{ramachandran17} or speeding up the convolutions by working in the frequency domain emloying Fourier transforms as in \cite{mathieu13}, \cite{rippel15}. Finally, it is well-known that correlations between data points are stronger on an intraday basis. Therefore, it might be interesting to test the model on intraday data to see if the ability of the model to learn long-term dependencies is even more valuable in that case. 


\section*{Acknowledgements}
This research is supported by the European Union in the the context of the H2020 EU Marie Curie Initial Training Network project named WAKEUPCALL. We also thank an anonymous referee for the constructive comments for improving the quality of the paper.

\bibliographystyle{siam}
\bibliography{biblio}
\end{document}